# Explainable Sentence-Level Sentiment Analysis for Amazon Product Reviews

Xuechun Li[+]
School of Mathematics and Statistics
Wuhan University
Wuhan, China
2018302010101@whu.edu.cn

Xueyao Sun[+]
Jinan University – University of Birmingham Joint Institute
Jinan University
Guangdong, China
xueyaosun828@gmail.com

Zewei Xu[+]
School of Mechanical Science and Engineering
Huazhong University of Science and Technology
Wuhan, China
604952495@qq.com

Yifan Zhou[+]
College of Arts and Sciences,
University of North Carolina at Chapel Hill
Chapel Hill, USA
yyifan@ad.unc.edu

+ They are all first authors

*Abstract*—In this paper, we conduct a sentence-level sentiment analysis on the product reviews from Amazon and thorough analysis on the model's interpretability. For the sentiment analysis task, we mainly use the Bi-LSTM model with attention mechanism. For the study of interpretability, we consider the attention weights distribution of single sentence and the attention weights of main aspect terms. The model has an accuracy of up to 96%. And we find that the aspect terms have the same or even more attention weights than the sentimental words in sentences.

*Keywords—sentiment analysis, product reviews, Bi-LSTM, attention, aspect terms*

I. INTRODUCTION

Nowadays, e-commerce and modern logistics are developing rapidly. One hallmark of this Internet era is that more people are switching their shopping preference to purchasing products online via e-commerce platform [1]. Online shopping is of significant convenience and efficiency. It is now popular with all age groups.

However, because online shopping enables only browsing virtual goods and one cannot physically see the product in person, many problems arise by the inconsistency between item description and products. To ensure that the product matches their description, customers use product reviews as an essential reference. Therefore, product reviews should be an essential index to evaluate commodities. Sentiment analysis, for example, can be applied to the reviews [2]. Both consumers and producers can be benefited from such quantified analysis.

Currently, there are several sentiment classification techniques, including document-level, sentence-level sentiment, aspect-level. There are also sub-tasks including multi-domain and multi-modal sentiment classifications [3].

For document-level, the essential unit is the document itself. The file is viewed as one whole element about a specific topic. For example, Onan [4] proposed the CNN-LSTM architecture consisted of five layers. Fauzi, M. Ali [5] studied the performance of Support Vector Machine (SVM) combined with TF-IDF, and Chen et al. [6] used the OPSM bi-clustering algorithm to identify the feature vectors.

The sentence-level sentiment classification uses sentence as the basic unit and extracts words such as adjectives and nouns, which expresses a particular polarity. For instance, the word 'amazing' has a prior positive polarity, and 'degrade' will be assigned with a prior negative polarity. If the word 'not' is used, however, the polarity will convert. For example, Smetanin and Komarov [7] used CNN and pre-trained Word2Vec for sentiment analysis.

The core task for aspect-level sentiment classification is the identification of opinions and aspects in a sentence. Take the phrases "Good keyboard, long battery life, largest hard drive and Windows 7" as an example. The aspect terms in these phrases are keyboard, battery life, hard drive and Windows 7. Opinion terms anticipated to be extracted are "good", "long" and "largest" [8]. Looking at another sentence example: "The size of the room was smaller than our expectation, but the view from the room would not make you disappointed." Based on the same principle, it is evident that the "room size" and "room view" are the aspects, and, respectively, they are negative and positive [9]. Aspect-level sentiment classification also can be applied widely.

Currently, many researchers focus on the three main tasks mentioned before. However, there is only few work exploring the interpretability of sentiment analysis models, i.e., the reason why the sentiment to be classified with the polarity [10]. In this paper, we follow the existing research, focus on sentence-level sentiment analysis and propose two ways to explore the reasons of the sentiment analysis results: top-$k$ check-worthy [11] words in one sentence and top-$k$ check-worthy aspect-terms [12] extracted from the whole dataset. Aspect-terms can reflect the rich information contained in a sentence, such as the main concerns [13]. Therefore, we propose to use the aspect-terms, i.e., the targets from the texts, to extract the users' sentiment towards specific features related to the products.

II. RELATED WORK

Briefly speaking, the most commonly used methods in this field are lexicon-based, machine learning based, and deep learning based methods. Many scholars have modified traditional machine learning methods and deep learning methods to fit the sentiment analysis task. The main modification is setting up the sentiment lexicon and yields approving results [14].

## A. Lexicon-Based Method

As is mentioned above, the most important modification is the sentiment lexicon construction. It involves calculating the sentiment from the semantic orientation of words or phrases that occur in a text. A dictionary of positive and negative words is required in this approach, with a sentiment polarity value assigned to each word. Broadly speaking, in lexicon-based approaches, a piece of text message is represented as a bag of words. Within this bag, the dictionary's sentiment values are assigned to all positive and negative words or phrases in the message.

## B. Machine Learning Based Method

Methods such as Word2Vec and FastText are able to extract word vector features from given texts. However, they are not capable of finding out the emotional features, which means that manual process is still required to assign the sentiment values. The most frequently used machine learning methods are Naive Bayes, Support Vector Machine, Maximum Entropy, Random Forest and Conditional Random Fields model [15] [16]. Beyond these, scientists have developed some more robust solutions.

## C. Deep Learning Based Method

Recently, deep learning methods are popular because of their relatively ideal performance in large-scale processing datasets. There are many popular models, including CNN, RNN, Bi-directional RNN, and, most recently, Attention-based networks. CNN is a popular category of neutral network model. It has shown significant success and innovation in computer vision and image processing [17]. Kim's modified CNN model to conduct sentiment classification is one of the most famous CNN models [18], which uses the combination of CNN and Word2Vec. This work shows that the high performance of pre-trained word embedding models can be a powerful tool to import CNN's performance. Li et al. [19] proposed a framework that combines different levels of prior knowledge into word embeddings for sentiment analysis. To further improve the embedding space structures, Yu et al. [20] combined sentiment lexicon and deep learning method to propose a model that utilize sentiment intensity scores to learn the embeddings. RNNs and the variants are the most frequently used deep learning models for processing sequential data. More specifically, a special type of RNN called Long Short Term Memory (LSTM) is showing its power in recent works [21] [22]. Chen et al. [23] discussed a divide-and-conquer approach for sentence-level sentiment classification using biLSTM-CRF. Yuan et al. [24] harnessed attention-based networks. Jiang et al. [25] and Zhang et al. [26] proposed an approach to combine model A and Model B target sentiment classification into a Q&A system using Attention-based GRU networks.

## III. DATASET

We use the dataset called Amazon Musical Instruments Reviews. This dataset is obtained from the Amazon platform, and it contains 10,261 pieces of product reviews. The dataset provides sufficient texts and many other factors that can be used to analyze. Fig. 1 shows the head of the Amazon Musical Instrument Review dataset.

TABLE I. DATASET OVERVIEW

| reviewerID | asin | reviewerName | helpful | reviewText | Overall summary | unixReviewTime | reviewTime |
|---|---|---|---|---|---|---|---|
| A2IBPI20UZIR0U | 13847192 | cassandra tu " Yeah, well, that's just like, u... | [0, 0] | Not much to write about here, … The product does exactly as it should and is quite affordable... | 5good | 1393545600 | 02 28, 2014 |
| A14VAT5EAX3D9S | 138471932 | Jake | [13, 14] | The primary job of this device is to block… | 5Jake | 1363392000 | 03 16, 2013 |
| A195EZSQDW3E21 | 138471932 | Rick Bennette " Rick Bennette" | [1, 1] | Nice windscreen protects my MXL mic and prevents pops... | 5It Does The Job Well | 1377648000 | 08 28, 2013 |
| A2C00NNG1ZQQG2 | 138471932 | RustyBill " Sunday Rocker" | [0, 0] | This pop filter is great. It looks... | 5GOOD WINDSCREEN FOR THE MONEY | 1392336000 | 02 14, 2014 |
| A94QU4C90B1AX | 138471932 | SEAN MASLANKA | [0, 0] | So good that I bought another one. Love… | 5No more pops when I record my vocals. | 1392940800 | 02 21, 2014 |
| A2A039TZMZHH9Y | B00004Y2UT | Bill Lewey "blewey" | [0, 0] | I have used monster cables for years, and … | 5The Best Cable | 1356048000 | 12 21, 2012 |
| A1UPZM995ZAH90 | B00004Y2UT | Brian | [0, 0] | I now use this cable to run from the output... | 5Monster Standard 100 - 21' Instrument Cable | 1390089600 | 01 19, 2014 |
| AJNFQI3YR6XJ5 | B00004Y2UT | Fender Guy "Rick" | [0, 0] | Perfect for my Epiphone Sheraton II. Monster cables are well constructed… | 5Didn't fit my 1996 Fender Strat... | 1353024000 | 11 16, 2012 |
| A3M1PLEYNDEYO8 | B00004Y2UT | G. Thomas "Tom" | [0, 0] | Monster makes the best cables and a lifetime warranty doesnt hurt either… | 5Great cable | 1215302400 | 07 6, 2008 |
| AMNTZU1YQN1TH | B00004Y2UT | Kurt Robair | [0, 0] | Monster makes a wide array of cables, including … | 5Best Instrument Cables On The Market | 1389139200 | 01 8, 2014 |
| A2NYK9KWFMJV4Y | B00004Y2UT | Mike Tarrani " Jazz Drummer" | [6, 6] | I got it to have it if I needed it. I have found that… | 5One of the best instrument cables within the brand | 1334793600 | 04 19, 2012 |
| A35QFI0M46LWO | B00005ML71 | Christopher C | [0, 0] | If you are not use to using… | 4It works great but I hardly use it. | 1398124800 | 04 22, 2014 |
| A2NIT6BKW11XJQ | B00005ML71 | Jai | [0, 0] | I love it, I used this for my... | 3HAS TO GET USE TO THE SIZE | 1384646400 | 11 17, 2013 |
| A1C0O09LOLVI39 | B00005ML71 | Michael | [0, 0] | I bought this to use in my home studio… | 5awesome | 1371340800 | 06 16, 2013 |
| A17SLR18TUMULM | B00005ML71 | Straydogger | [0, 0] | I bought this to use with my keyboard... | 5It works! | 1356912000 | 12 31, 2012 |
| A2PD27UKAD3Q00 | B00005ML71 | Wilhelmina Zeitgeist "coolartsybabe" | [0, 0] | This Fender cable is the perfect length for me!... | 2Definitely Not For The Seasoned Piano Player | 1376697600 | 08 17, 2013 |
| AKSFZ4G1AXYFC | B000068NSX | C.E. "Frank" | [0, 0] | | 4Durable Instrument Cable | 1376352000 | 08 13, 2013 |

## IV. METHODOLOGY

The methodology (Fig. 1) consists of four steps used in this research.

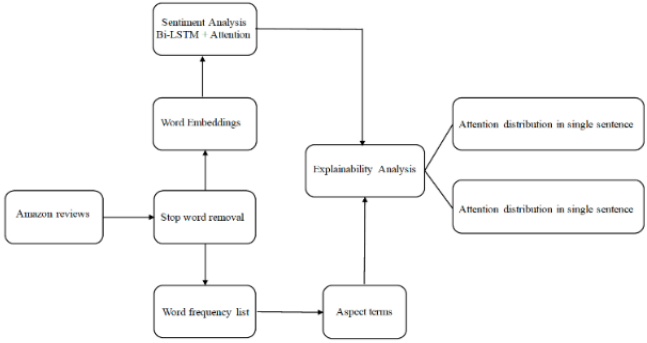

Fig. 1. Methodology

### A. Data Pre-processing

We take the Amazon Musical Instruments Reviews as an example to illustrate our data pre-processing. First, we utilized the columns named "reviewText" and "overall" from the dataset as the target. Then we delete the analphabetic sign, lowercase all words and remove the meaningless words such as pronouns and conjunctions using the "stopwords" from the Python NLTK API.

### B. Constructing Aspect-Term Set

We compute each word's word frequency by applying TF-IDF vectorizer from python package "scikit-learn" [27] and sort it in decreasing order. Then we extract the top 160 frequency nouns as the aspect-term set. Notice that in the aspect-term set, all the words are directly obtained from the original texts. Thus, we avoid the manual work of labelling the aspect.

### C. Set up the Model

*1) Model Description*
The overall structure is illustrated in Fig. 3

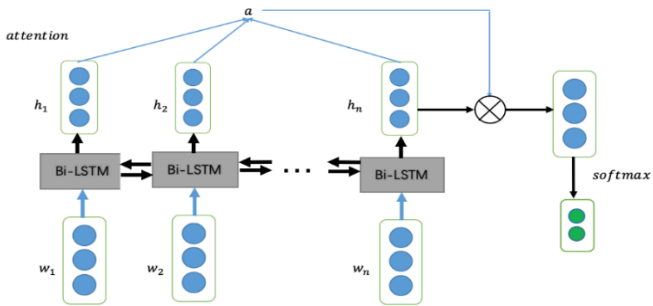

Fig. 2. Model Structure

*2) Construct sentiment lexicon*
The function of the sentiment lexicon is to give each word $w_i$ in $S$ a corresponding sentimental weight $Sw_i$ [28]. There are several widely used sentiment dictionaries such as the General Inquirer (GI), the NRC Emotion Lexicon and VADER include, and the popular automatically generated sentiment lexicons include SentiWordNet (SWN), Sentiment 140 lexicon, and NRC Hashtag Sentiment Lexicon [29].

In this paper, we choose WordNet and SentiWordNet as the dictionaries. We determined that for some words, the sentiment results differ between SentiWordNet and WordNet. To overcome the opposing scores, we normalize the sentiment by taking the average of the two scores. For example, for the word 'love', SentiWordNet gives a weight of 0.71, while WordNet gives 0.0 as the sentiment weight. Therefore, we propose to use the average weight from the two dictionaries to generate more solid sentiment weights. Notice that different part of speech of the same word will be assigned with different weight in both dictionaries. Hence in the aim of consistency, we take the mean of all the weights of a word as its final weight.

$$senti(w_i) = \begin{cases} sw_i & w_i \text{ in } SD \\ 1 & w_i \text{ not in } SD \end{cases}$$

where $SD$ is the sentiment lexicon, and $senti(w_i)$ is the sentiment weight for $w_i$.

*3) Word embedding*
Word2Vec [30] is one of the most popular techniques for learning word embeddings. It yields vector outputs of words. Similarly, Global Vectors(GloVe) [31] generates the vector encoding of a word. Additionally, char2vec [32] can learn embedding related to each character of a word. Other popular models include ELMo [33] and BERT [34]. In this paper, we use BERT to generate word embeddings.

In BERT, every word will be converted to a word vector $v_i$, and the sentiment weight is used to weigh the word vector.

$$x_i = v_i * senti(w_i)$$

*4) Bi-LSTM layer*
One method to extract the features from text is to use the Recurrent Neural Network (RNN) model. It uses the previous output and the next word together as input, generates the following output, and learns the sequential feature dynamically [35]. RNN is capable of figuring out the correlation between the front and the back sequence. However, the model faces several drawbacks, especially the gradient disappearance problem [36]. Because of this disadvantage, the length of sequential data generated by RNN is limited. The hidden layer of the RNN can be modified, which in turn improves the LSTM.

The first layer in the LSTM model is the forget gate $f$. It determines the information to forget in memory cell at the last moment [37]. The function is defined as:

$$f_t = \sigma(W_f x_t + U_f h_{t-1} + V_f c_{t-1} + b_f) \quad (1)$$

where the parameters stand for:

$h_{t-1}, x_t$: inputs of LSTM unit,

$c_{t-1}$: the state of memory cell at the last moment,

$W.$: the connecting weight of $x_t$ and $\cdot$,

$U.$: the connecting weight of $h_{t-1}$ and $\cdot$,

$V.$: the connecting weight of $c_{t-1}$ and $\cdot$,

$b.$: the bias term,

and $\sigma(\cdot) = \frac{1}{1+e^{-x}}$: the sigmoid activation function.

The next layer is the input gate $i$ which determines the information to be updated in memory cell at current time [38]. The calculation is given by:

$$\begin{cases} i_t = \sigma(W_i x_t + U_i h_{t-1} + V_i c_{t-1} + b_i) \\ c\_in_t = tanh(W_c x_t + U_c h_{t-1} + V_c c_{t-1} + b_c) \\ c_t = f_t \cdot c_{t-1} + i_t \cdot c_i n_t \end{cases} \quad (2)$$

The last layer is the output gate $o$ [39]. It generates the output to be used in the attention layer.

$$\begin{cases} o_t = \sigma(W_o x_t + U_o h_{t-1} + V_o c_{t-1} + b_o) \\ h_t = o_t \cdot \tanh(c_t) \end{cases} \quad (3)$$

In our paper, we adopt the Bi-LSTM model. It introduces feature of Bi-RNN into the LSTM model so that it calculates the information bidirectionally [40]. The information can be propagated both forward and backward.

*5) Attention layer*

Given a piece of text, each word in the text will play a different role in the sentence. The words have different sentiment polarity, and thus influence the sentiment direction of the sentence [41]. Some words do not affect the sentence sentiment, while some significantly contribute to the sentiment polarity. Thus, we should adopt a mechanism to assign different weights to the words in a sentence.

In our paper, we use the self-attention mechanism to generate the weights. For the hidden state $h_i$ of the Bi-LSTM output, the attention weight $a_i$ is given as:

$$u_i = \tan(W \cdot h_i + b) \quad (4)$$

$$a_i = \frac{e^{u_i^T \cdot u_w}}{\Sigma_i e^{u_i^T \cdot u_w}} \quad (5)$$

where $W$ is the weight matrix, $b$ is the offset, and $u_w$ is the global context vector.

Then, the hidden layer output $h_i$ is weighted by $a_i$ and the weights are put together in a feature vector $S$.

$$S = (s_i) = (a_i \cdot h_i) \quad (6)$$

*6) Loss function*

To calculate the accuracy of our model, we introduce a loss function. There are several different functions that are usually chosen as loss function [42]. Here we combine the Mean Square Error (MSE) function with Cross Entropy (CSE) function to obtain the loss.

$$Loss(i) = \frac{1}{2} MSE(i) + \frac{1}{2} CSE(i) \quad (7)$$

where,

$$MSE(i) = \sqrt{\frac{1}{N} \Sigma_i (y_i - f(x_i))^2} \quad (8)$$

$$CSE(i) = -\frac{1}{N} \Sigma_i y_i \log(f(x_i)) \quad (9)$$

*D. Interpretability Analysis*

*1) Attention distribution in single sentence*

In the attention layer described, we obtain the attention weights for each word in every sentence. Attention weights can describe the importance of words in one sentence. In this part, we formulate attention distribution in single sentence by applying $softmax$ to the attention weights

*2) Aspect-term Weights Analysis*

In this part, we compute the attention weight of each word in the aspect-term set:

$$aw_i = f(\{attnw_i^j\}) \quad (10)$$

where $aw_i$ represents the attention weight for the $i$-th aspect term, $attnw_i^j$ represents the attention weights for the $i$-th aspect term in the $j$-th sentence, and $f$ can be chosen in $\{max, sum, average\}$.

## V. EXPERIMENT

In this section, we input the dataset to evaluate our model. We modify the hyperparameters in the model and test their effect on model performance. By these adjustments, we select proper hyperparameters and obtain the following results.

*A. Dataset Description*

We have mentioned the six datasets used in our experiment. Here we will use the Amazon Musical Instrument Reviews as the example dataset to illustrate our experiment process. Fig. 4 below illustrates the overall sentence length distribution.

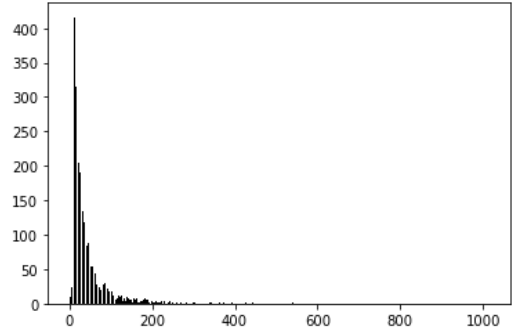

Fig. 3. Sentence length

*B. Vectorization Process*

We use Word2Vec to convert the text information into vectors. The parameters set in this word embedding step is displayed in Table 2.

TABLE II. WORD2VEC PARAMETERS

| Parameter | Description | Value |
|---|---|---|
| size | length of the vector | 200 |
| alpha | initial learning rate | 0.025 |
| min_count | word frequency lower than this will be omitted | 5 |
| max_vocab_size | maximum value of the vocabulary list | None |
| iter | numbers of iteration | 5 |
| batch_word | | 10000 |

*C. Evaluation Indicators*

We define the evaluation indicators as an index to evaluate the model.

First of all, we define the calculation parameters:

(1) TP (true positive): the number of positive comments classified correctly;

(2) FP (false positive): the number of negative comments classified wrongly;
(3) TN (true negative): the number of negative comments classified correctly;
(4) FN (false negative): the number of positive comments classified wrongly.

With these four classifications, the evaluation indicator can be defined as:

(1) Accuracy: the ratio of comments that are classified correctly;

$$accuracy = \frac{TP+TN}{TP+FP+TN+FN} \quad (11)$$

(2) Precision: the ratio of positive comments labelled correctly to all positively labelled comments;

$$precision = \frac{TP}{TP+FP} \quad (12)$$

(3) Recall: the ratio of negative comments labelled correcly to all negatively labelled comments;

$$recall = \frac{TN}{TN+FN} \quad (13)$$

(4) $F_1$ score: the weighted average of precision and recall.

$$F_1 = \frac{2 \cdot precision \cdot recall}{precision+recall} \quad (14)$$

*D. Hyperparameters*

We divide our dataset into two parts: 70% is the training set, and the rest 30% is the test set. The four hyperparameters mainly studied are epochs, batch size and dropout rate. We adjust these three hyperparameters respectively and study their effect on the performance of the model.

*1) Epochs*

Epochs means how many iterations are conducted in the training set [43]. The generalization ability of the model increases if significant epochs are set. However, we cannot set the epochs too large because this will lead to overfitting, which reduces the generalization ability [44]. Hence it is crucial to adjust different epochs values and select one with the best performance.

TABLE III. EPOCH ADJUSTMENT

| Epoch | Accuracy | Precision | Recall | F1 |
|---|---|---|---|---|
| 8 | 96.0% | 96.0% | 99.9% | 97.9% |
| 10 | 94.6% | 94.6% | 99.9% | 97.2% |
| 12 | 95.0% | 95.1% | 99.9% | 97.5% |
| 17 | 95.7% | 95.7% | 99.9% | 97.8% |

Table 3 illustrates how the performance score will change by different epochs. It can be seen that with the growth of epochs, the scores decrease and then increase again. When epoch is 8, the model obtains the best performance in all evaluation metrics. Therefore, we select epoch to be 8.

*2) Batch Size*

Batch size is the size of data that are input into the Bi-LSTM model at one time [45]. Generally speaking, a larger batch size will improve model efficiency and accuracy. However, due to the limitation of computer memory, we should select a proper batch size to balance model accuracy and computer performance.

TABLE IV. BATCH SIZE ADJUSTMENT

| Batch size | Accuracy | Precision | Recall | F1 |
|---|---|---|---|---|
| 32 | 95.3% | 95.3% | 99.9% | 97.6% |
| 34 | 95.2% | 95.2% | 99.9% | 97.5% |
| 36 | 95.5% | 95.5% | 99.9% | 97.7% |
| 38 | 96.0% | 96.0% | 99.9% | 97.8% |

Table 4 illustrates how the four evaluation metrics change with batch size. From the table, we can see that the recall scores are almost the same among all batch sizes. However, when batch size increases, accuracy, precision and F1 score oscillate in a small range. Although a batch size of 38 gives the best model performance, since it occupies too much computer memory, we select 32 as the final batch size.

*3) Dropout Rate*

The dropout rate is the percentage of cells that are removed from the network. Dropping out cells enables cells in the hidden layer to randomly select other cells to learn features. It prevents hidden layers from coadaptation [46]. However, if the dropout rate is too high, there will be too much information loss, leading to underfitting. In our training process, the dropout parameter enables the model to randomly stop training some of the cells and forget the features of comments they contain.

TABLE V. DROPOUT RATE ADJUSTMENT

| Dropout | Accuracy | Precision | Recall | F1 |
|---|---|---|---|---|
| 0.2 | 95.4% | 95.6% | 99.8% | 97.7% |
| 0.4 | 95.8% | 96.0% | 99.8% | 97.8% |
| 0.6 | 95.2% | 95.5% | 99.9% | 97.5% |
| 0.8 | 94.7% | 95.4% | 99.2% | 97.3% |

Table 5 illustrates how evaluation metrics change with dropout rate. From the table we can see that, as dropout rate increases, the accuracy first increase and then decrease, and so does recall and F1. When dropout rate is 0.4, all the evaluation indexes reach maximum. Therefore, we select the dropout rate to be 0.4

The final model performance with parameters given above are shown in Table 6.

TABLE VI. DROPOUT RATE ADJUSTMENT

| Accuracy | Precision | Recall | F1 |
|---|---|---|---|
| 96.0% | 96.0% | 99.9% | 97.9% |

*E. Comparison with Baseline*

In this part, we compare the performance of our model with the baselines: SVM, Naïve Bayes, and LSTM. The result is shown in the table below (Table 7). Based on the implementation from scikit-learn.org, we choose 'rbf' kernel for SVM, and alpha equals to 1 for Naïve Bayes.

TABLE VII. DROPOUT RATE ADJUSTMENT

| | Accuracy | Precision | Recall | F1 |
|---|---|---|---|---|
| SVM | / | 48.0% | 50.0% | 49.0% |
| Naïve Bayes | 84.4% | 87.9% | 84.7% | 83.6% |
| LSTM | 95.1% | 95.2% | 99.9% | 97.5% |

*F. Interpretability Analysis*

*1) Attention weight distribution in single sentence*

To illustrate the distribution of the attention weights, we draw the thermography of review 1 in Fig. 4. Note that the content of review 1 is "fine cable decent price point nothing exceptional mind gets job done well enough" after being processed by stopwords. We find that the words "fine", "price", "job", "enough" have higher attention weights, which is in accordance with commonsense that the sentiment word and the most concerned aspects such as "price" will have higher attention. Fig. 5 gives the attention weight for each word in review 1

| fine | cable | decent | price | point | nothing | exceptional | mind | gets | job | done | well | enough |
|------|-------|--------|-------|-------|---------|-------------|------|------|------|------|------|--------|
| 0.24 | 0.097 | 0.0097 | 0.12  | 0.00035 | 5.00E-05 | 5.00E-05 | 0.0072 | 0.089 | 0.18 | 0.014 | 0.095 | 0.15 |
| 1 | 2 | 3 | 4 | 5 | 6 | 7 | 8 | 9 | 10 | 11 | 12 | 13 |

Fig. 4. Attention weight distribution in review 1

*2) Aspect terms attention weight analysis*

In this part, we extract the top 160 words with the highest TF-IDF weights to construct the aspect terms set and use $f = mean$ to obtain their attention weights. Fig. 4 shows the highest aspect terms' attention weights. Fig. 5 shows the attention weights of the highly sentimental words. We can see that, on average, the aspect terms have higher attention weights than the sentimental words. The aspect term 'fender' has the highest attention weights, indicating that it is the feature of musical instruments that users care about most. Fig. 6 shows the aspect terms with top 10 aspect weights. Fig. 7 shows the sentimental words with top 9 attention weights.

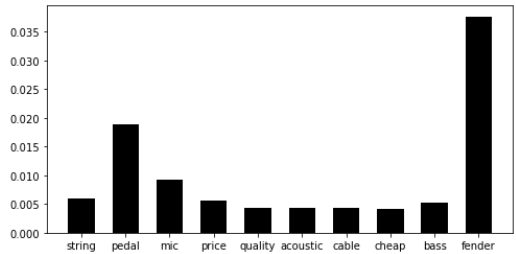

Fig. 5. Aspect terms with high attention weights

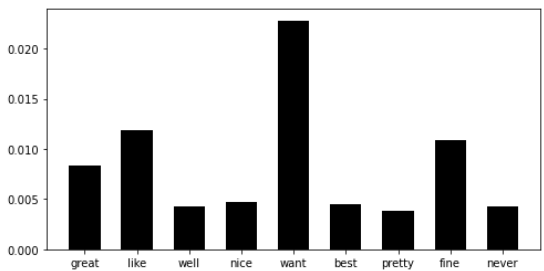

Fig. 6. Sentimental words with high attention weights

## VI. CONCLUSION

In the rapid development of online shopping, a sentiment analysis technology of product reviews has gained more and more attention. Most customers will have a solid first impression from the product reviews on a large proportion of online shopping platforms. Therefore, sentiment analysis plays an increasingly important role in helping the sellers determine whether the products are satisfying and helping the customers get a better profile of the products. In this paper, we use Bi-LSTM and attention mechanism to conduct sentiment analysis.

As we can see in the experiment session, the model achieved excellent results. Also, based on the attention weights obtained from the attention layer, we find that on average, the main aspect terms such as 'fender' and 'pedal' have bigger attention weights than the highly sentimental words such as 'great', 'nice', i.e., the users' sentiment to the products is based mainly on their sentiment to the main aspect terms. Therefore, we believe using these weights can help the sellers better determine the aspects of the products that need improvement. We noticed that the recall value for the Amazon Product Reviews of Musical Instruments is exceptionally high during our experiments. Since a high recall value means there are very few false negative and that the classifier is more permissive in the criteria for classifying something as positive, we conclude that for the biased datasets, which means that there is a vast difference between the number of positive and negative samples, the recall value(if the negative samples are few) or the precision value(if the positive samples are few) will be extraordinarily high and lost the test meaning. We also find that most of the existing deep learning methods only analyze unbiased datasets through our investigation. Therefore, for future work, we will continue to optimize the model to better work on biased datasets.

ACKNOWLEDGEMENT

Every author's contribution to the paper is the same. The ranking is in alphabetical order.